\documentclass[sageapa,times]{sagej}

\usepackage{moreverb,url}

\usepackage[colorlinks,bookmarksopen,bookmarksnumbered,citecolor=blue,urlcolor=blue,linkcolor=blue]{hyperref}

\newcommand\BibTeX{{\rmfamily B\kern-.05em \textsc{i\kern-.025em b}\kern-.08em
T\kern-.1667em\lower.7ex\hbox{E}\kern-.125emX}}

\def\volumeyear{2023}

\let\bibhang\relax
\usepackage[natbibapa]{apacite} 
\let\cite\citep

\usepackage[T1]{fontenc}
\usepackage{graphicx}
\usepackage{subcaption}
\usepackage{booktabs}
\usepackage{multirow}
\usepackage{amsmath}
\usepackage{enumitem}
\usepackage{listings}
\usepackage{xcolor}

\begin{document}

\title{Framing Analysis of Health-Related Narratives: Conspiracy versus Mainstream Media}

\runninghead{Reiter-Haas et al.}

\author{Markus Reiter-Haas\affilnum{1}, Beate Kl\"{o}sch\affilnum{2}, Markus Hadler\affilnum{2}, and Elisabeth Lex\affilnum{1}}

\affiliation{\affilnum{1}Institute of Interactive Systems and Data Science, Graz University of Technology\\
\affilnum{2}Department of Sociology, University of Graz}

\begin{abstract}
Understanding how online media frame issues is crucial due to their impact on public opinion. Research on framing using natural language processing techniques mainly focuses on specific content features in messages and neglects their narrative elements. Also, the distinction between framing in different sources remains an understudied problem. We address those issues and investigate how the framing of health-related topics, such as COVID-19 and other diseases, differs between conspiracy and mainstream websites. We incorporate narrative information into the framing analysis by introducing a novel frame extraction approach based on semantic graphs. We find that health-related narratives in conspiracy media are predominantly framed in terms of beliefs, while mainstream media tend to present them in terms of science. We hope our work offers new ways for a more nuanced frame analysis.

\end{abstract}


\maketitle
\clearpage

\section{Introduction}
\label{sec:intro}

\begin{figure}[t]
    \centering
    \captionsetup{type=figure,belowskip=-6pt}
    \includegraphics[width=0.75\columnwidth]{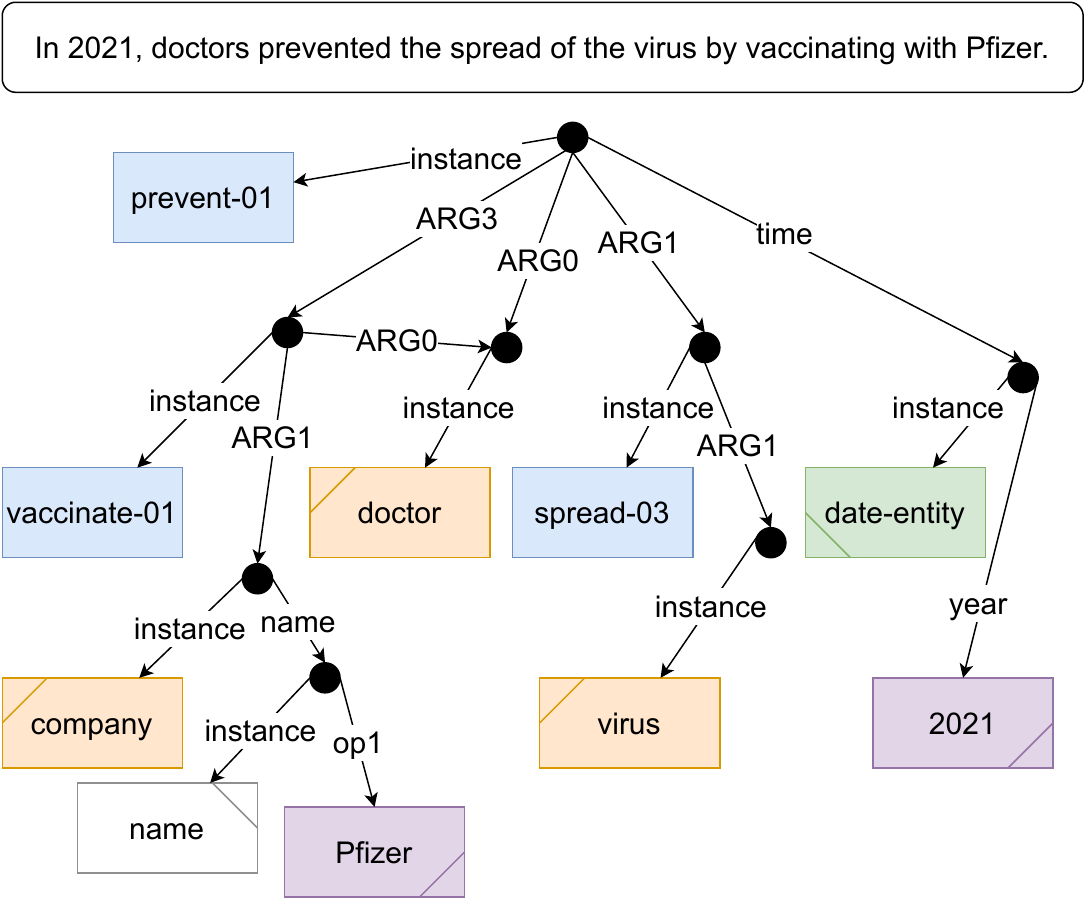}
    \caption{Example sentence (top) with its extracted AMR graph using a BART-based model. Given this representation, we can identify the narrative elements, while syntactical information such as tenses is omitted. Within the narrative, three characters are present, i.e., a \emph{doctor} who acts twice (i.e., two \emph{ARG0} relations) as character (orange), as well as a \emph{company} and a \emph{virus} (both with \emph{ARG1} relations). The plot (blue; predicates with word senses) revolves around three frames, namely \emph{prevent}, \emph{vaccinate}, and \emph{spread}. 
    Additionally, the year 2021 (i.e., \emph{date-entity} = setting; green) and the company \emph{name} Pfizer are depicted as entities (purple). 
    }
    \label{fig:graph}
\end{figure}

The perception of reality in human communication heavily relies on how messages are framed, leading to significant effects on human behavior. In their groundbreaking study, \citet{tversky1985framing} demonstrate that altering the formulation of a problem impacts people's decision making. Hence, understanding the role of framing in textual communication is a critical research direction. While many existing computational framing research studies have primarily focused on narrow topics, such as war~\cite{wicke2020framing}, terrorists~\cite{demszky2019analyzing}, morality~\cite{reiter2021studying}, or blame~\cite{shurafa2020political}, these works often overlook the narrative content embedded within the frames. However, understanding the narrative content is essential as it plays an important role in transmitting the underlying information. Furthermore, framing in textual content is defined as promoting particular aspects of information through the selection and salience of content~\cite{entman1993framing}. Hence, a comprehensive framing analysis needs to extend beyond the identification of frames themselves and interpret why frames were used, such as supporting a particular narrative.

Narrative framing is especially critical in domains where the differences in the messages can be subtle, as exemplified in conspiracy theories. According to \citet{mcleod2022navigating}, narrative frames are abstract constructs that refer to entire messages rather than individual content features, which makes them difficult to identify.
\citet{fong2021language} find that despite conflicting narratives of conspiracy theories versus scientific narratives, the language employed within them has similarities in terms of word frequencies~\cite[using LIWC - Linguistic Inquiry and Word Count;][]{tausczik2010psychological}. Contrary to \citeauthor{fong2021language} hypotheses, these similarities also extend to causal (e.g., “how,” “why,” “because”) and outgroup language (e.g., “they,” “them”). 
Still, the authors highlight consistent differences in specific linguistic patterns, such as the prevalent use of anger-based wording to convey negative emotion within conspiratorial discourse. 
In our work, we aim to bridge the gap between word-frequency of messages and their framing. To that end, we propose to incorporate narrative information into the framing analysis of conspiracy theories. 

We argue that most conventional text analysis methods employed in framing research, including various types of topic modeling \cite[see][for an overview]{ali2022survey} do not accurately capture narrative information.
Therefore, we propose the extraction of \textbf{semantic graphs} from textual data to conduct frame analysis. Our approach draws on the recent work of \citet{jing2021characterizing}, in which semantic relations are mined from textual data in the form of triples, i.e., subject, predicate, and object. In this representation, subjects and objects correspond to semantic roles such as agents and patients, while the predicate (i.e., the \emph{verb}) connects these roles. In our work, however, we leverage semantic graphs as they allow for a more comprehensive representation of concepts, entities, and relations in textual data. This enables us to capture a broader range of semantic information beyond roles and predicates.   

To that end, we utilize abstract meaning representations~\cite[AMR;][]{banarescu2013abstract} as a means to transform textual content into semantic graphs. AMR gives us a structured representation of textual content in the form of AMR graphs, from which we extract their inherent semantic frames using edge information; thus considering the embedded narrative content contained within them.
To validate the effectiveness of our approach, we apply it to health-related narratives mined from the language of a conspiracy corpus~\cite[LOCO;][]{miani2021loco}. LOCO contains text documents from mainstream and conspiracy websites from a time period of 2004 until 2020. 
The study of \citet{miani2021loco} shows that the detection of conspiracy content compared to mainstream content can be challenging for humans, primarily due to ambiguity, highlighting the potential benefits of algorithmic support tools for humans.
Our work aims to address this challenge by showing the difference in narrative elements, such as setting, characters, plot, and the moral of the story. Figure~\ref{fig:graph} exemplarily shows an extracted AMR graph on COVID-19 and its narrative elements. It depicts how narrative elements can be determined by edge traversal of the given semantic information. This approach allows for a more effective analysis of narratives, as compared to traditional methods relying on syntactical or word-based extraction techniques, which would prove considerably more challenging in this context.

Through our analysis, we uncover a distinctive pattern in the narrative framing employed by conspiracy media compared to mainstream media. Specifically, conspiracy media tend to employ belief-based rather than science-based arguments. 
Conversely, mainstream media shows the opposite tendency (i.e.,  towards science rather than beliefs). This disparity in narrative framing underpins the contrasting approaches to information dissemination between both media types. Hence, our approach advances the narrative understanding of textual content by providing a comprehensive and holistic view of embedded narrations. Consequently, our methodology enables a nuanced frame extraction, facilitating future works in framing applications.

In sum, our main contributions are:     
\begin{itemize}[noitemsep, nosep]
    \item[\emph{C1}]:~We present a novel approach based on AMR and use it to extract frames imbued with narrative information. Our approach is flexible while also being conceptually simple to employ.
    \item[\emph{C2}]:~We demonstrate that within LOCO, the framing of health-related narratives (i.e., on COVID-19, diseases in general, and pharmacology) of conspiracy media focuses on beliefs compared to mainstream media, which focuses on science. 
\end{itemize}

\section{Background}
\label{sec:background}

In this section, we provide the background of narrative framing analysis. We describe the theory behind narrations, review the related work of computational frame extraction, and introduce the background of AMR, i.e., frame semantics.

\paragraph{Narration.}
According to \citet{piper-etal-2021-narrative}, narrations contain multiple elements, such as a setting (in the work of Piper et al., setting refers to spatial location exclusively; in contrast, our work 
also considers temporality), 
characters (referred to as agents and potential objects), a plot (referred to as events), a reason (which we refer to as moral of the story), as well as a perspective (i.e., information about the teller and recipient). 
In this work, we focus mainly on the content and consider the perspective implicitly by contrasting two different sources. In line with \citeauthor{entman1993framing}'s~\citeyearpar{entman1993framing} basic assumptions, we assume that the tellers within each source have different motivations.
Similar to \citeauthor{piper-etal-2021-narrative}'s definition of narration, the narrative policy framework~\cite{jones2010narrative}, on which we ground our work, defines a set of four narrative elements. It comprises a setting or context, a plot, the characters, and the moral of the story. In this framework, the perspective is also implicit by considering the trust and credibility of the source (i.e., narrator). 

In our work, we leverage \emph{abstract meaning representations (AMR)}~\cite{banarescu2013abstract} to extract frames. AMR is a graph-based approach to representing the semantic content of textual information. AMR parsing transforms textual information into a directed acyclic graph, whose nodes correspond to concepts~\cite{xu2020improving}. These concepts are connected via edges reflecting semantic relations, such as, e.g., the role that they occupy.
AMR parsers are trained on an annotated corpus consisting of structured semantic information, which is based on a strict specification of how AMR graphs are constructed by humans~\cite{banarescu2012abstract}. We use AMR because it results in simple representations; also, it allows us to extract frames in a flexible, exploratory manner.

\paragraph{Computational frame extraction.}
A body of recent research focuses on computational frame extraction~\cite[see][for an overview]{ali2022survey}. Herein, framing detection is often presented as a classification problem, such as in the SemEval 2023 shared task~\cite{semeval2023task3}.
In another example, \citet{tourni-etal-2021-detecting-frames} consider gun violence as portrayed in news headlines and lead images. In their work, the authors formalize the notion of frame concreteness derived from the tangibility of words within their headlines. They relate it to the relevance of images to the given headline. Their experiments show that news about politics has a high concreteness and relevance, whereas news about society/culture is low on both. \citet{huguet-cabot-etal-2020-pragmatics} consider the frames security and defense, morality and fairness, and equality in the context of immigration, gun control, and death penalty. Their approach is based on RoBERTa~\cite{liu2019roberta}, and they predict the framing of policy issues based on a joint model of emotion, metaphors, and political rhetoric. From a methodological perspective, many works also employ unsupervised learning to extract frames, such as clustering and sentiment analysis~\cite[e.g.,][uses both]{burscher2016frames}. Herein, \citet{jing2021characterizing} extract frames in partisan tweets related to COVID-19 by combining BERT with semantic role labeling~\cite{shi2019simple}. 

We similarly position our paper as frame extraction but focus on narration instead.
In our work, we extract frames related to COVID-19, albeit from conspiracy and mainstream media. 
Different from previous works~\cite[e.g.,][]{jing2021characterizing}, we use AMR instead of topic modeling or semantic role labeling\endnote{We also experimented with a BERT variant for both topic modeling and semantic role labeling but found richer AMR representations better suited for the task at hand.}. By structuring concepts within a text rather than tagging text spans, AMR allows for more flexible extraction of semantic information, which in turn benefits the interpretability of the data. For instance, extracted roles (i.e., agents and patients) are often long sequences of text in semantic role labeling, as modifiers are also included in the tagged spans of text. Hence, the full sequence of \emph{the doctors who only recently graduated} is tagged as an agent instead of just extracting the \emph{doctor} concept.

\paragraph{Frame Semantics.}

Frame semantics has a long history in the natural language processing community since its initial introduction by \citet{fillmore1976frame}. The use of frame semantics gained momentum with the FrameNet project~\cite{baker1998berkeley} and aided their adoption in natural language understanding~\cite{fillmore2001frame}. PropBank~\cite[short for proposition bank;][]{palmer2005proposition} provided an annotated corpus of frames and the relations to their arguments, and hence, 
paved the way for widely used computational methods in natural language understanding, such as semantic role labeling~\cite{shi2019simple} and AMR parsing~\cite{banarescu2013abstract}. Our work uses the latter due to its matching properties for the task.

As shown in Figure~\ref{fig:graph}, 
AMR is a graph-based representation of the semantic content in the text without explicit syntax. 
To be more concise, AMR is a rooted, directed, acyclic graph with labeled edges and leaf nodes.
AMR parsing converts texts to structured information beyond the capabilities of simple textual extraction methods. Firstly, it enriches the textual information with semantic information about data types (e.g., \emph{date-entity}) and 
information (e.g., \emph{name}). It also simplifies the semantic information by normalization (e.g., removing tenses -- \emph{prevented} to \emph{prevent}, singularizing nouns -- \emph{doctors} to \emph{doctor}, considering word senses -- \emph{spread-03} to indicate distribution instead of smearing, omitting the distinction between nouns and verbs -- converting both \emph{vaccinating} and \emph{vaccination} to the common form \emph{vaccinate-03}, and even substituting for named entities -- \emph{company} instead of using its name \emph{Pfizer}).

\section{Notation and Definitions}
\label{sec:terminology}
As the term "frame" is used in various ways in the literature (e.g., compare \citet{entman1993framing} and \citet{fillmore1976frame}), we briefly clarify at this point the specific meaning of the underlying representation and most important terms for the remainder of the paper. 
Our definitions are adapted to be specific for the \emph{framing analysis imbued with narrative information}.

As an alternative to the graph-based representation, AMR graphs can also be represented as serialized text using the Penman notation~\cite{kasper-1989-flexible}. The Penman notation applies to connected, rooted, directed, acyclic, and labeled graphs, such as AMR, which is often even used synonymously~\cite{goodman2019amr}. The notation has a recursive structure concerning its relations denoted by parenthesis, typically also indicated using newlines and indentation as a convention for human readability. 
As an example, Figure~\ref{fig:graph} is equivalent to the Penman notation in Listing~\ref{lst:penman}.
\lstset{caption={Penman Notation of AMR graph},label=lst:penman}
\begin{lstlisting}[float,floatplacement=H]
    (p / prevent-01
      :ARG0 (d / doctor)
      :ARG1 (s / spread-03
            :ARG1 (v / virus))
      :ARG3 (v2 / vaccinate-01
            :ARG0 d
            :ARG1 (c / company
                  :name (n / name
                        :op1 "Pfizer")))
      :time (d2 / date-entity
            :year 2021))
\end{lstlisting}

In the following, we will clarify the definitions using the AMR annotation guideline~\cite{banarescu2012abstract} and the provided example. 
We highlight the \textbf{main building blocks} in bold, the references to the \emph{Penman example} in italic, and 'lexical definitions' with single quotation marks.

\textbf{Semantic frames} are defined in a language resource (here, PropBank\endnote{\url{https://propbank.github.io/v3.4.0/frames/}}), which comprises a predefined set of \textbf{predicates} including their \textbf{sense} and associated \textbf{frame arguments}. 
For instance, consider the first two lines in Listing~\ref{lst:penman}. The frame \emph{(p / prevent-01 :ARG0 (d / doctor))} comprises a predicate (\emph{prevent-01}), with \emph{doctor} as frame argument. 
Frame arguments have a \textbf{semantic role} assigned to them (e.g., \emph{ARG0} for \emph{doctor}). In the given example, the \emph{prevent} predicate only has a single sense (i.e., \emph{$01$} with the meaning of 'stopping in advance'). However, when considering the next two lines (\emph{:ARG1 (s / spread-03 :ARG1 (v / virus))}), \emph{spread-03} refers to spread in the third sense - 'cause to be widely located or distributed' rather than referring to 'smear' (i.e., spread-01) or 'extend' (i.e., spread-02). Also, frame arguments can themselves be frames, with \emph{spread-03} being an argument of \emph{prevent-01} as denoted by the \emph{ARG1} relation. Therefore, frames can contain substructures, such as \emph{virus} belonging to \emph{spread-03} as argument. Alternatively, frame arguments can be \textbf{concepts} comprising words and phrases (e.g., \emph{doctor} or \emph{virus}).
Furthermore, Penman uses variables (equivalent to nodes in the graph) to distinguish between their \textbf{instances} and denote instance relations with a slash. Hence, in the example, the \emph{d} variable of the semantic frame \emph{(v2 / vaccinate-01 :ARG0 d)} refers to the same \emph{doctor} as in the \emph{(p / prevent-01 :ARG0 (d / doctor))}. Finally, nodes can have associated \textbf{attributes} (e.g., \emph{company} has the \emph{name} attribute of \emph{"Pfizer"}).

Here, we want to emphasize the subtle difference between semantic frames (also called "linguistic frames") and narrative framing (i.e., a form of communicative frames), which operate on different levels of language and communication, respectively~\cite{sullivan2023three}\endnote{Both, in turn, rely on a third type of cognitive frames, which operate at the level of thought. While implicitly required, cognitive frame are not the focus in the present work and thus omitted.}. In the task at hand, language is essential for studying narrative framing, and therefore depend on semantic frames (which is not necessarily the case for other types of communicative frames, such as art~\cite{sullivan2023three}). Consequently, we use the precisely defined semantic frames as a basis to study more complex communicative frames (i.e., narrative frames). 

When considering the narrative information in the framing analyses, i.e., \textbf{narrative framing}, we refer to AMR-subgraphs as potential narratives, such as \emph{(p / prevent-01 :ARG0 (d / doctor) :ARG1 (s / spread-03 :ARG1 (v / virus)))}, and instances, as well as attributes, as \textbf{narrative elements}. In the remainder of the paper, we refer to semantic frames as \emph{frames}, frame arguments as \emph{arguments}, and narrative elements as \emph{elements}.
For brevity, we omit semantic roles and use subject-verb-object notation where applicable (e.g., \emph{doctor prevent-01 spread-03}).

\section{Method}
\label{sec:method}

We present our approach for frame mining in text-based content based on AMR, %
comprising a pipeline of three main components (i.e., contribution~\emph{C1}):
\begin{enumerate}[label=\arabic*.]
    \item \emph{AMR parsing} with a pretrained BART model and Penman decoder.
    \item\emph{Mining narrative elements}, such as characters, plot, setting, and the moral of the story.
    \item \emph{Analysis of narrative information} concerning differences in word usage, embedding spaces, and subgraphs. %
\end{enumerate}
We first describe the main components in detail, before providing a complete conceptual \emph{description of the pipeline} from a technical perspective.

\subsection{AMR Parsing}
\label{sec:parsing}
For AMR parsing, i.e., the conversion from text to AMR graphs, we use the AMRlib\endnote{\url{https://github.com/bjascob/amrlib}} with a pretrained BART-based model~\cite[i.e., \emph{parse\_xfm\_bart\_base-v0\_1\_0}; based on ][]{lewis2019bart}. The model was trained on the AMR Annotation Release 3.0~\cite[LDC2020T02;][]{knight2021abstract} based on the PropBank annotations~\cite{palmer2005proposition} and has a SMATCH score of $82.3$, which is a semantic matching score based on F1-measure~\cite[refer to][for details]{cai2013smatch}. 
AMR parsing also has the advantage of applying multiple linguistic tasks simultaneously, such as co-reference resolutions via reentrants in the graph~\cite[refer to][for an overview of different types of reentrants]{szubert-etal-2020-role} and named entity recognition (via the name attribute). Hence, it alleviates the need for building sophisticated processing pipelines. The output of the AMR parser is in PENMAN notation, which is transformed into a graph for mining via the Penman library~\cite{goodman-2020-penman}.

\subsection{Mining Narrative Elements}
\label{sec:mining}
We first introduce the narrative policy framework~\cite{jones2010narrative}, which describes an empirical approach to studying policy narratives. Thereby, a narrative structure consists of characters (e.g., heroes/villains or victims), a plot (i.e., actions), a setting or context, as well as the moral of the story. 
The narrative policy framework provides the theoretical grounding for mining the narrative information. Specifically, we extract the narrative elements, i.e., characters and plots, by considering the AMR edge information. 

Characters and plots are described as simple (<subject>, <verb/predicate>, <object>) triples, such as, e.g., \emph{we protect them}. A more general representation of the plot and its corresponding characters is as a variable-length tuple of the format: (<predicate>, <argument0>, <argument1>, \ldots, <argumentN>), which resembles PropBank frames. Frame arguments can be other frames (e.g., vaccinate-01 or spread-03), concepts (e.g., nouns such as doctor, company, or virus), or attributes (e.g., named entities such as Pfizer or a year such as 2021).

\paragraph{Characters.}
For the characters (orange), we consider instances of \emph{ARG0} or \emph{ARG1} roles. While frames can have more than these two arguments (i.e., \emph{ARG2} and beyond), they tend to appear less often and hence play a less important role, as the highest-ranked (i.e., the lowest number) argument precedes according to the PropBank guidelines~\cite{babko2005propbank}. Hence, we focus on the first two arguments for simplicity. Due to reentrants in the graph, characters can assume multiple (possibly even different) roles. This is exemplified in Figure~\ref{fig:graph} as seen by the doctor in the example, who acts twice as a character. 

According to the narrative policy framework, characters can be categorized as heroes, villains, or victims. In the present work, we do not distinguish between these subtypes of characters\endnote{A naive approach to model the subtypes, is to use sentiment analysis to distinguish between heroes (positive) and villains (negative) portrayed characters. However, sentiment is not part of AMR (only sentence polarity) and thus would require external resources (e.g., dictionaries or models). As our work focuses on AMR for narrative analysis, we omit such analysis for brevity and leave it as future work.}.

\paragraph{Plot.}
For the plot (blue), we use the predicates of the semantic frames directly. To find the \emph{frames} (i.e., predicates), we reverse the traversal of the graph (i.e., go up from \emph{ARG0} or \emph{ARG1} arguments to parent nodes and towards their instances). One observation is that the plot is driven by verbs and indicated by other words that can be encoded in frames. In the example given in Figure~\ref{fig:graph}, the spread is part of the plot, as it suggests the distribution of a virus (i.e., \emph{ARG1}) but does not detail who the spreader is (i.e., misses an \emph{ARG0}).

\paragraph{Setting.}
For the setting (green), we consider the special \emph{time} and \emph{location} relations.
These represent the context 
in which the narration is embedded and are typically associated with attributes (purple), such the specific \emph{year}. 
Compared to the characters and plot, the attributes can be more diverse as they are not bounded by the number of common words and their normalization. For instance, considering the range of pharmaceutical companies researching vaccinations for COVID-19, some associated named entities are more commonly portrayed (e.g., Pfizer), while others are only rarely mentioned (e.g., Sanofi). 
Similarly, certain temporal or spatial information might appear more frequently in relation to particular topics (e.g., 2021 for COVID-19). Nevertheless, these types of information are unbounded by definition. 
In the analysis, we thus differentiate between types of the settings, such as the narrative refer to a \emph{year}, and their specific attributes, e.g., 2021.

\paragraph{Moral of the Story.}
Similar to the setting, the moral of the story (i.e., reason) relies on specific relations, i.e., \emph{purpose} and \emph{cause}. Unlike the setting, these relations often comprise concepts or even complete subgraphs. Here, we use the top element (i.e., root of the subgraph), which carries the most meaningful information. Moreover, as many sentences do neither include a purpose nor cause, such relations are only sparsely available. Nevertheless, they provide important narrative information.

\subsection{Analysis of Narrative Information}
\label{sec:overrep}
We compare the narrative information extracted between the mainstream and conspiracy corpus. Herein, we use the log-odds ratio to diminish the influence of predominant characters and plot devices in terms of relative frequency to each other, e.g., similar to ~\citet{jing2021characterizing}. However, we leverage smoothed log-odds ratio instead of informative Dirichlet priors~\cite{monroe2008fightin}, and thus do not require a separate background corpus. The complete equation, which also includes Z-score normalization, is given by:

\begin{equation}
\label{eq:zscore_logodds}
    z_w = \frac{log\frac{f_i(w)+1}{n_i-f_i(w)+1} - log\frac{f_j(w)+1}{n_j-f_j(w)+1}}{\sqrt{\frac{1}{f_i(w)+1}+\frac{1}{f_j(w)+1}}}
\end{equation}

, where $f_i(w)$ and $f_j(w)$ represent the frequency of a given word $w$ in its corresponding sub-corpus, while $n_i$ and $n_j$ represent the total number of words per sub-corpus (i.e., $n_i=\sum_{w\in V}f_i(w)$ with $V$ containing all words and similarly for $n_j=\sum_{w\in V}f_j(w)$). Hence, the enumerator of Equation~\ref{eq:zscore_logodds} corresponds to a relative probability that is symmetric due to the log transformation, while the denominator accounts for the variance. Consequently, over-represented words in the given sub-corpus get a high absolute value. The sign indicates the dominant sub-corpus, while the magnitude of the score (i.e., absolute value) is equivalent for both sub-corpora.
Hence, negative values show the over-representativeness in the alternative sub-corpus without requiring recalculation.

\paragraph{Visualization for elements.}
We plot the over-represented words (indicating plot, characters, setting, and moral of the story) in a shared two-dimensional embedding space using UMAP-reduced embeddings~\cite{mcinnes2018umap} of the model's input layer side by side for comparison\endnote{We also experimented with PCA for dimensionality reduction and pretrained GloVe embeddings, which are two other often used approaches, respectively. However, we find that UMAP better preserve semantic similarity, while using the model's inherent embeddings allows for a better mapping, as it avoids a domain shift.}. The positioning of the plot improves the analysis by positioning semantically similar words in a similar region, and thus improves the subsequent interpretation. 
For readability, we use a force-adjusted positioning for the labels. To declutter the plot, we simplify the labels by removing the sense tags (as a distinction between word senses is typically not necessary anyway in this particular case). In a similar vein, we only keep the first part of compound concepts, e.g., \emph{government} instead of \emph{government-organization}, which follows the same rationale.

\paragraph{Notation for narratives.}
For readability, we also provide a short notation to represent frames and their corresponding arguments, as well as their associated z-score.
We denote the frames with \emph{ARG0} $\xleftarrow{1.0}$ \emph{FRAME-01} and \emph{FRAME-01} $\xrightarrow{1.0}$ \emph{ARG1} respectively. Specifically, \emph{ARG0} appears left of the frame with a left arrow, while \emph{ARG1} appears on the right with a right arrow. Consequently, the two relations can be combined to form an ARG0--FRAME--ARG1 triplet. For instance, \emph{doctor} $\xleftarrow{1.0}$ \emph{prevent-01} $\xrightarrow{1.0}$ \emph{spread-03} reads similar to the well-known subject--verb--object structure. 
Above the arrows, we provide the z-score of the log-odds-ratio between the two corpora.

\subsection{Pipeline Description}
\label{sec:detail}
The text is tokenized and fed into an embedding layer of a pretrained BART model. The input token embeddings are combined with positional embeddings and fed into a bidirectional encoder stack comprising multiple encoder layers for text understanding. The resulting representation is then in turn fed into an autoregressive decoder stack (again comprising multiple decoder layers) to iteratively generate the PENMAN representation. A PENMAN decoder then creates a graph-based representation. By traversing the graph from its root, the Frame Miner component extract the relevant information (narrative elements). The aggregation of the information is then divided depending on the label. Using the frequency information, we can compare the occurrences and calculate a score over-representative elements for each label. We use the top-N (positive score) and bottom-N (negative score) elements and plot them in a word embedding space for analysis. Here, we want to stress that we distinguish between different word types, such as Frame and ARG0. Note that we also provide a detailed diagram of the approach in the supplemental materials. %

Additionally, we emphasize that the approach is easily extensible. For instance, we could inject sentiment information to distinguish between the usage of words from the word frequencies, i.e., to derive the character sentiment for a hero vs. villain distinction. Similar, other information could be extracted by including dictionaries, e.g., for a value-based analysis. However, this goes beyond the scope of work, i.e., pure AMR-based analysis of narrative information.

\section{Experiments and Results}
\label{sec:results}
We present our analysis and empirical results of health-related framing (i.e., contribution~\emph{C2}). Specifically, we investigate health-related narratives and report our findings in three topics (i.e., Covid-19, general diseases, and pharmacology). To that end, we leverage a publicly available dataset (i.e., LOCO) containing media content from various online information sources.

\subsection*{Dataset and Preprocessing}
\label{sec:dataset}
\addcontentsline{toc}{subsection}{Dataset and Preprocessing}

We use the LOCO dataset~\cite{miani2021loco}, which contains documents collected from English-speaking news websites concerning both mainstream and conspiracy media\endnote{In a pre-study, we analyzed whether similar topics are discussed in the mainstream and conspiracy corpus using BERTopic~\cite{grootendorst2022bertopic}. We observed differences, especially regarding the discussed nouns, as conspiracy media are more concerned with topics such as COVID-19 origin, vaccination, and President Trump. In contrast, mainstream media focuses on drug trials, testing, and the economy. Hence, such a method is too limited to analyze narratives as it gives us mainly the context of a story; we, however, are interested in the characters, plot, and the moral of the story.}.
The documents were collected from May to July 2020 via web scraping (thus, also including older documents dating back to 2004 for the oldest conspiracy document) using a combination of predefined sources and manual seed selection while excluding non-English domains from the collection (refer to \citet{miani2021loco} for the complete data collection and processing details).
Documents are labeled as \emph{conspiracy} if they originate from a website known to publish "unverifiable information that is not always supported by evidence" (as determined by the Media Bias/Fact Check list\endnote{\url{https://mediabiasfactcheck.com/conspiracy/}}) and mainstream otherwise.  The corpus comprises $72,806$ mainstream documents from $92$ websites and $23,937$ conspiracy documents from $58$ websites on $47$ seeds.

We consider three health-related subcorpora. 
First, we focus on the documents on COVID-19-related topics, i.e., we use the following seeds as defined by LOCO: \emph{vaccine.covid}, \emph{covid.19}, and \emph{coronavirus}. Second, we consider documents related to disease with the seeds \emph{aids}, \emph{cancer}, \emph{zika.virus}, and \emph{ebola}. Third, the pharmacology LOCO subset comprises documents with the seeds \emph{vaccine}, \emph{pharma}, and \emph{drug}.

\begin{figure}[t]
    \centering
    \captionsetup{type=figure, justification=justified,}
    \begin{subfigure}[b]{0.47\textwidth}
        \includegraphics[width=0.95\columnwidth]{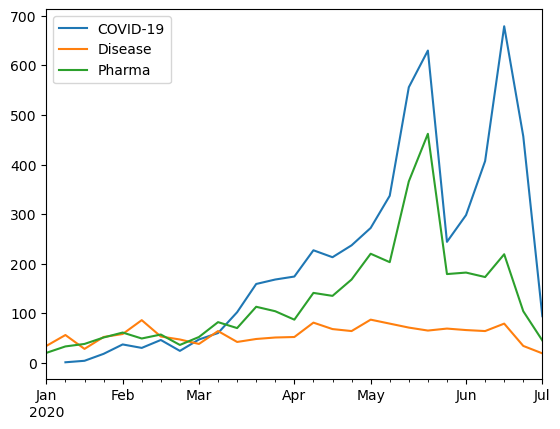}
        \caption{Number of documents in LOCO per subcorpus since 2020. We observe an increasing trend for COVID-19 and Pharmacology, with two separate peaks in May and June. Hence, it resembles the COVID-19 waves in English-speaking countries~\cite{owidcoronavirus}. The relative peaks are about equal for COVID-19, while the first peak is far more pronounced in Pharmacology. No such trend is present for the disease dataset. We omitted plotting the data before 2020 due to a lack of noteworthy peaks.}
        \label{fig:doc_time}
    \end{subfigure}
    \hfill
    \begin{subfigure}[b]{0.47\textwidth}
        \includegraphics[width=0.95\columnwidth, trim={0 0.3cm 0 0},clip]{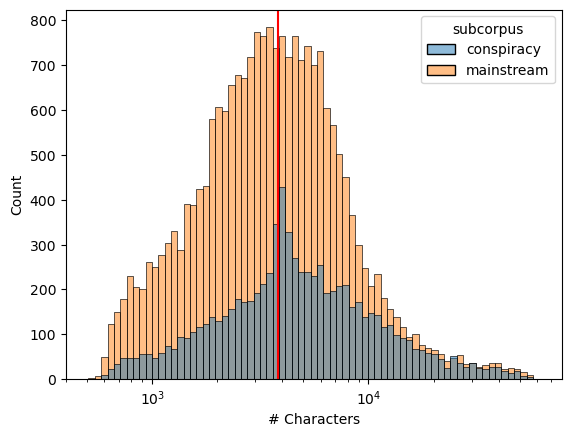}
        \caption{Histogram of document length, i.e., number of characters, in the three subcorpora of LOCO. The X-axis is on a log scale. Hence, both conspiracy and mainstream media resemble lognormal distributions, but conspiracy media is more pronounced in the median (as depicted by the red line). This observation is consistent both when considering individual subcorpora, and concerning the number of words and sentences. Overall, there are more mainstream than conspiracy documents. 
}
        \label{fig:distribution}
    \end{subfigure}
    
    \caption{Details of the LOCO dataset in terms of temporality and distribution.}
    \label{fig:dataset}
\end{figure}

Considering the time (see Figure~\ref{fig:doc_time}), we observe that the majority of documents for COVID-19 and pharmacology appear in 2020 with peaks in May and June (COVID-19 specific peak) just before the end of data collection on July 3\textsuperscript{rd}, 2020. Note, however, that both disease and pharmacology have more documents overall compared to COVID-19, which is in turn more clumped in 2020. This in turn also result in a greater number of graphs and narrative elements.

In Figure~\ref{fig:distribution}, we observe that both mainstream and conspiracy media resemble a lognormal distribution in terms of document length (we only depict the distribution for the number of characters, but observe similar distributions for the number of words and sentences). On average, each document consists of $5455$ characters, $1009$ words, and $38$ sentences. However, conspiracy documents are more concentrated near the median (i.e., red line at $3805$). 
We extract the AMR graphs using the methodology described in the \nameref{sec:method} Section\endnote{We ran the calculation on a shared SLURM-managed server using a single Nvidia Quadro RTX 8000. The calculation for each of the three subsets took approximately a day but could differ depending on server utilization.}. The detailed statistics are described in Table~\ref{tab:num_elements}.

\begin{table}[t]
    \centering
    \resizebox{\columnwidth}{!}{%
\begin{tabular}{l|rr|rr|rr}
{} & \multicolumn{2}{c|}{COVID-19} & \multicolumn{2}{c|}{Disease} & \multicolumn{2}{c}{Pharma} \\
\multicolumn{1}{c|}{total \#} & conspiracy & mainstream & conspiracy & mainstream & conspiracy & mainstream \\
\midrule
Documents       &      2,414 &      6,308 &      2,877 &      8,296 &      3,914 &      9,839 \\
Graphs     &     99,728 &    255,622 &    150,189 &    287,897 &    215,294 &    364,979 \\ \hline
Plots &    436,780 &  1,193,282 &    622,873 &  1,332,290 &    921,102 &  1,758,036 \\
$\hookrightarrow$ unique &  6,577 & 7,656   &  7,573 & 8,439  &     8,303 & 9,398   \\
\rule{0pt}{3ex}%
Characters &   419,764    &   1,143,711  & 598,708     &  1,270,719 &  882,955   &  1,689,006   \\
$\hookrightarrow$ unique &  12,500    &  15,981  & 15,354  & 17,890  &  16,845   &  20,177 \\
\rule{0pt}{3ex}%
Settings   &     68,807 &    195,564 &     96,141 &    221,733 &    128,206 &    248,783 \\
$\hookrightarrow$ unique &    3,042 & 4,243  &  3,861 & 4,720  &     4,103 & 4,999  \\
\rule{0pt}{3ex}%
Moral o.t.S.    &      9,433 &     25,899 &     12,562 &     25,772 &     19,022 &     38,179 \\
$\hookrightarrow$ unique &    1,769 & 2,371   &  2,069 & 2,339 &     2,438 & 2,945   \\
\rule{0pt}{3ex}%
Entities   &    164,993 &    395,068 &    250,144 &    510,859 &    341,763 &    596,920 \\
$\hookrightarrow$ unique &    17,339 & 36,379   &  27,102 & 40,754  &     30,613 & 48,659   \\
\bottomrule
\end{tabular}
}
    \caption{Dataset statistics regarding the number of extracted elements. Each document contains several graphs, which in turn contains elements of different types. We also report the number of unique elements per type.}
    \label{tab:num_elements}
\end{table}

\begin{figure*}[t]
    \centering
    \captionsetup{type=figure, justification=justified,}
    \begin{subfigure}[b]{0.99\textwidth}
        \frame{
        \includegraphics[width=.99\columnwidth]{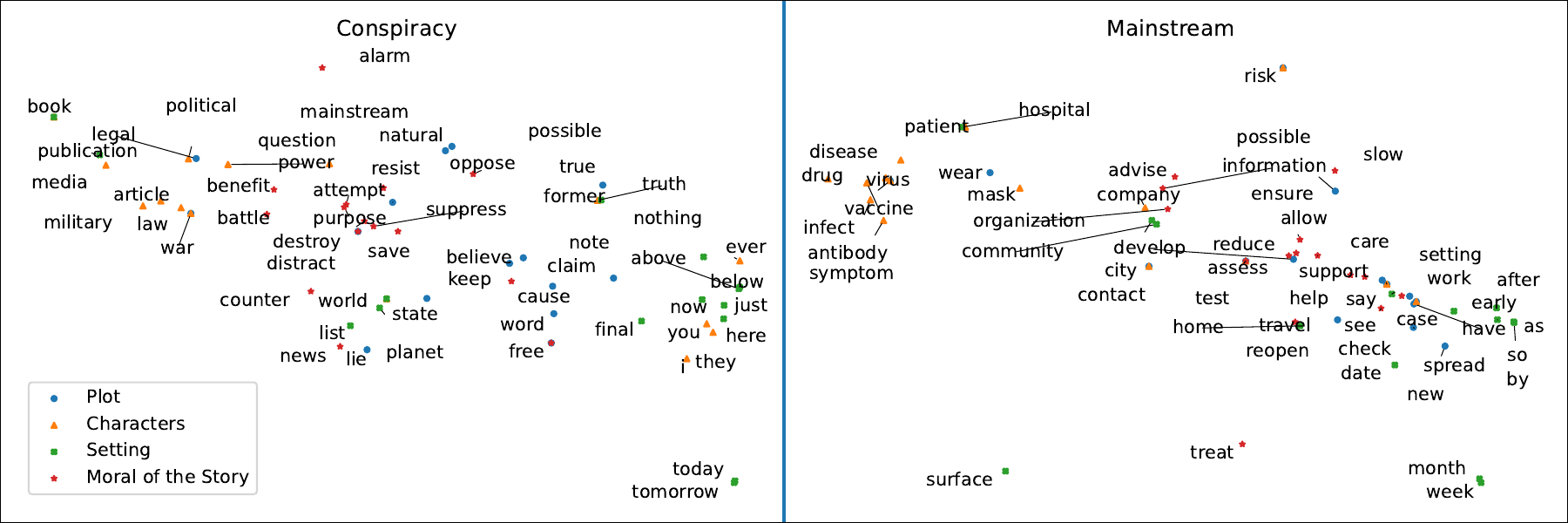}
        }
        \caption{COVID-19}
        \label{fig:ccombined}
    \end{subfigure}
    
    \begin{subfigure}[b]{0.99\textwidth}
        \frame{
        \includegraphics[width=0.99\columnwidth]{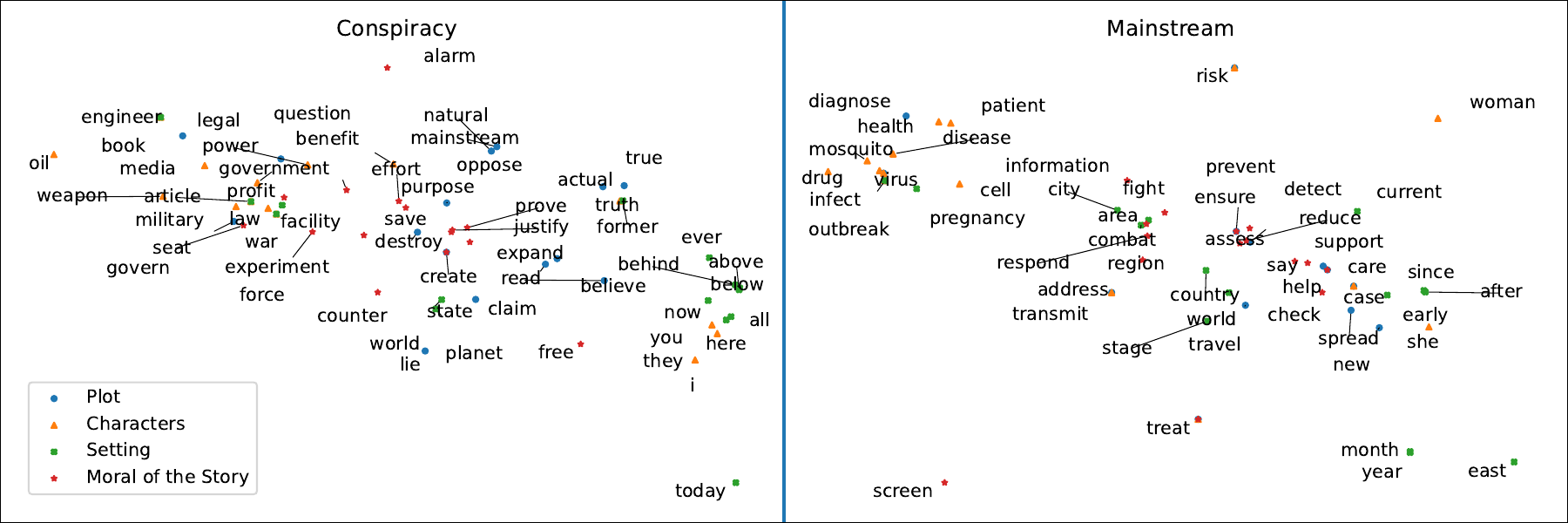}
        }
        \caption{Diseases}
        \label{fig:dcombined}
    \end{subfigure}
    
    \begin{subfigure}[b]{0.99\textwidth}
        \frame{
        \includegraphics[width=0.99\columnwidth]{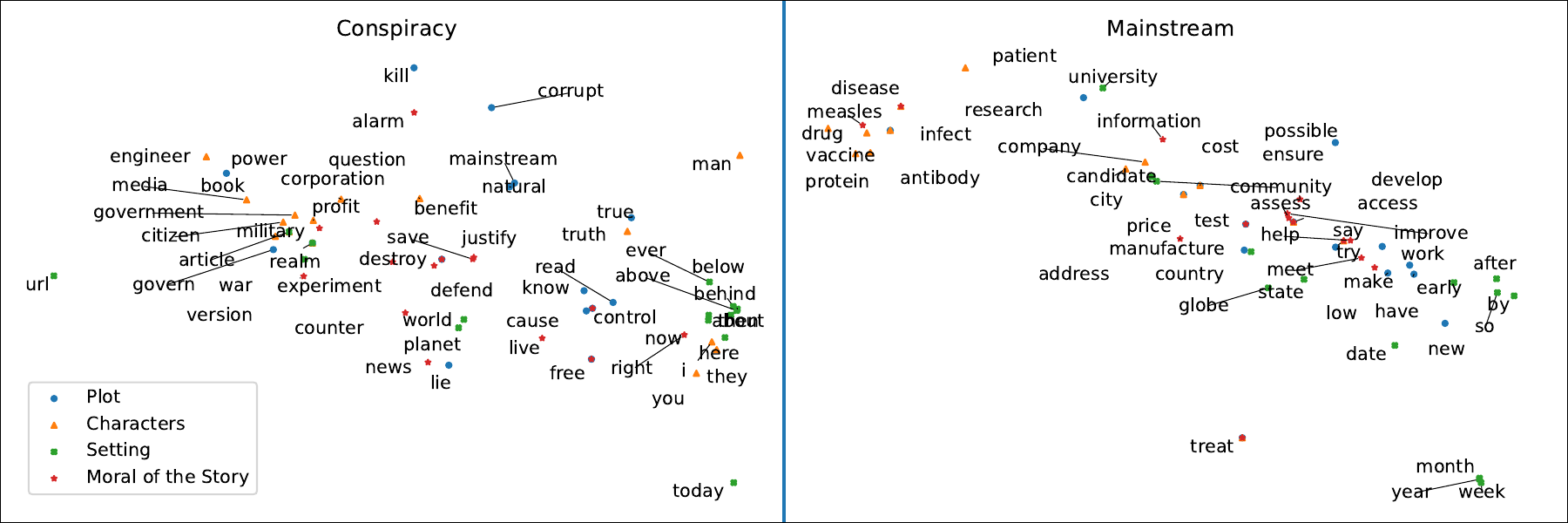}
        }
        \caption{Pharmacology}
        \label{fig:pcombined}
    \end{subfigure}
    
    \caption{Over-represented \textbf{narrative elements} (i.e., plot, characters, setting, moral of the story) on COVID-19 in conspiracy versus mainstream media. Positioning is according to 2-dimensional UMAP embedding of the AMR input layer (i.e., semantically similar words appear in similar locations), and labels are force-adjusted for readability (with lines indicating their associated positioning if moved beyond a threshold).} 
    \label{fig:combined}
\end{figure*}

\subsection*{Analysis of Narrative Information}
\label{sec:analysis}
\addcontentsline{toc}{subsection}{Analysis of Narrative Information}
We contrast the narrative framing of COVID-19 in the mainstream and conspiracy corpus. In Figure~\ref{fig:ccombined} (we also provide the Table with the top 15 over-represented words per corpus per narrative element type with their associated score in the supplementary materials.), we observe that conspiracy media tends to focus on argumentation frames as plot, such as \emph{believe}, \emph{claim}, \emph{lie}, and \emph{oppose}. Conversely, mainstream media focuses on action-oriented frames like \emph{develop}, \emph{spread}, and \emph{reopen}. 
Similarly, mainstream media uses science-related characters such as \emph{scientists}, \emph{vaccines}, \emph{antibodies}, and \emph{proteins}. In comparison, conspiracy media use typical characters that suggest large-scale conspiracies, such as \emph{world}, \emph{elites}, \emph{truth}, and  \emph{power}. 
When considering the contexts, we note that conspiracy media is more focused on the \emph{now}, such as \emph{today} or \emph{tomorrow}, rather than specific \emph{weeks} or \emph{months} as is the case for mainstream media.
Finally, when considering the moral of the story, conspiracy media reasons more concerning \emph{alarm}, while mainstream uses \emph{information} in its narratives. 

Another difference is the war-focused framing in conspiracy media (e.g., using \emph{destroy} as frame, \emph{military} as character, and \emph{counter} as moral of the story). Whereas mainstream media has a more health-oriented framing (e.g., \emph{infect} being used both for the plot and as character, while \emph{treat} acts as rationale).
Besides, we also briefly investigated the associated attributes (see Table in supplementary materials\endnote{As attributes can be arbitrary, such as names of entities, their embeddings cannot directly be extracted from the AMR model. Hence, they do not possess a specific position in the graph, which is why we omitted plotting them and refer to the data instead.}), such as named entities, where we observe that narratives in conspiracy media revolve about people, such as \emph{Gates} and \emph{Trump}, as well as religion (e.g., \emph{Jews} and \emph{Christians}) and have a focus on \emph{US}/\emph{China}. In comparison, mainstream media focus on institutions, such as \emph{universities} and the \emph{NHS}.  

In the disease dataset (see Figure~\ref{fig:dcombined}), we observe many similarities to the COVID-19 dataset. However, we also notice a shift, especially in the entities of mainstream media, toward global south countries where the diseases are more prevalent. Furthermore, in conspiracy media, the narrations shift toward non-natural origins such as \emph{engineering}, \emph{weapons}, and \emph{chemical}. 

In the pharmacology dataset (see Figure~\ref{fig:pcombined}), the mainstream media uses the drug company names as entities and the \emph{development} and \emph{manufacturing} as plots with treatment-related characters such as \emph{dose}. Conspiracy media shows its mistrust with terms like \emph{corrupt}, \emph{kill}, and \emph{control}.

While all three datasets exhibit similarities, we also observe specific differences. Most notably, the mainstream disease dataset has a stronger emphasis on the role of women due to female-associated elements (e.g., \emph{she}, \emph{woman}, \emph{pregnancy}, \emph{care}).

\paragraph{Analysis of Narratives.}
To gain a clearer picture of how the frames are used, we investigate the differences in arguments (i.e., \emph{ARG0} and \emph{ARG1}) in three frames from the initial example (i.e., \emph{prevent-01}, \emph{spread-03}, and \emph{vaccinate-01}).
In general, we find that \emph{ARG1} is more suitable for the frames, as they have the highest scores. Here, we highlight noteworthy examples of narratives.

In COVID-19, conspiracy media mainly invokes \emph{prevent-01} $\xrightarrow{3.7}$ \emph{violence}, but also invokes the \emph{government-organization} $\xleftarrow{3.6}$ \emph{prevent-01} $\xrightarrow{2.9}$ \emph{individual}. Hence, their focus does not lie in the prevention of the virus. In comparison, mainstream media focus on the infection with \emph{prevent-01} $\xrightarrow{5.1}$ \emph{infect-01}. For \emph{spread-03}, conspiracy theories often focus on spreading rumors but also invoke \emph{vaccine} $\xrightarrow{4.2}$ \emph{spread-03}, suggesting that the vaccine spreads the disease. In contrast, mainstream media has a clear focus on the viral spread with \emph{person} $\xleftarrow{1.0}$ \emph{spread-03} $\xrightarrow{3.3}$ \emph{virus}. For \emph{vaccinate-01}, \emph{military} $\xleftarrow{2.9}$ \emph{vaccinate-01} is common for conspiracy media, whereas, \emph{vaccinate-01} $\xrightarrow{1.8}$ \emph{person} is common for mainstream media. We also analyzed differences in frame arguments. As a noteworthy example, conspiracy media is less concerned about preventing the virus and that the vaccine might spread the disease.

We observe similar patterns for diseases in general and pharmacology. 
Regarding the usage of the \emph{prevent-01} frame in pharmacology, we observe \emph{person} $\xleftarrow{4.3}$ \emph{prevent-01} in conspiracy and \emph{prevent-01} $\xrightarrow{6.8}$ \emph{infect-01} in mainstream media as the top (i.e., over-representative) narratives. Similarly, the usage of \emph{spread-03} frame in other diseases, \emph{vaccine} $\xleftarrow{4.3}$ \emph{spread-03} and \emph{spread-03} $\xrightarrow{4.1}$ \emph{virus} are dominant for conspiracy and mainstream media, respectively. Hence, the narratives are mostly mirrored between the different sub-datasets.

\section{Discussion}
\label{sec:discuss}
We now briefly discuss the implications of our findings on five distinct aspects. 

\paragraph{Narrative Themes.}
Apart from the well-known belief and faith focus of conspiracy outlets, our analyses of the different narratives highlights that the conspiracy sites emphasize an urgency of the social problem ("Today", "Now", etc.) and also an immediacy of the issue ("You", "I", and "We"). 
These characterizations are in line with the description of conspiracy adherer~\cite{douglas2019understanding}. 
Interestingly, conspiracy sites also see the mainstream and media as characters and part of the game, whereas mainstream does not refer to conspiracists - which points to a non-reciprocity. 
Similarly, we find a corroborative emphasis ("truth", "true", "actual"), which might suggest another demarcation to mainstream media.
Finally, while war-framing is often present in health-related discourse~\cite[e.g., as shown in ][]{wicke2020framing}, we observe a one-sided tendency towards war-framing in conspiracy media.

\paragraph{Societal Implications.}

Our work shows a clear distinction between the narratives in conspiracy and mainstream media. While this finding on its own is expected, we can draw parallels to prior studies. For instance, \citet{shelton2020post} suggests that the COVID-19 pandemic was the first \emph{post-truth pandemic}. Our work complements this finding, as a belief-oriented framing in conspiracy media competes with a science-oriented framing in mainstream media. 
Thus, we see that the media have different lines of argumentation on the same issue, which can influence and bias people's attitudes and drive polarization, e.g., through diverging assessments of the consensus in society regarding the seriousness of the COVID-19 pandemic \cite{logemann2023media}. A deeper understanding of the main competing narratives is therefore essential for improving the ability to spot fake news \cite{porshnev2021effects} and for automatic detection to identify and combat conspiracy theories in the media \cite{shahsavari2020conspiracy}. Knowledge of conspiracy narratives can thus be used to improve the dissemination guidelines of social media platforms or handbooks developed by policymakers, e.g., the 'Check Before You Share Toolkit' in the UK \cite{bloomfield2021communicating}. It can also be used to educate society at an individual level, e.g., through 'fake news games' where players learn to identify manipulation techniques commonly used in conspiracy theories \cite{basol2021towards}.
Yet, we need to acknowledge that our analysis is only highlighting the structure of arguments, but does not consider the reach of these sources.
However, based on previous research~\cite{reiter2022polarization}, we can estimate that COVID-19 conspiracy beliefs are more widespread than belief in other conspiracy theories~\cite{uscinski2022have}. 
Assessing 
content and arguments in online media is thus of utmost importance. 

\paragraph{Methodological Advancements.}
Our work is based on the premise that
text analyses are challenging, especially when considering more abstract concepts, such as framing (partly also due to a lack of a clear definition \cite{entman1993framing}). Hence, graph representations such as AMR allow for a more comprehensive analysis, which is supported by our approach. Most text processing tasks are directly handled by AMR parsing, which is conceptually easy to employ using pretrained models based on the Transformer architecture~\cite{vaswani2017attention}. Our approach demonstrates that narrative elements can directly be mapped onto AMR and thus extracted. 
For instance, the four elements of the narrative policy framework~\cite{jones2010narrative} are directly applicable. Moreover, similar elements as described by \citet{piper-etal-2021-narrative} can be extracted (besides the perspective, as it is typically not part of the textual content). Finally, we show that we can perform a wide range of analyses on the extracted information. 

\paragraph{Technical Limitations.}
We recognize two main limitations of our work: 
First, while AMR provides an expressive semantic representation of narrations, more subtle information, such as sentiment, cannot be extracted directly. %
Hence, 
AMR graphs would require external resources for sentiment analysis (e.g., sentiment polarity lexicons); moreover, while AMR encodes direct negations, considering indirect negations (e.g., via \emph{prevent-01} frame) would require yet another external resource. Second, while we show that AMR provides understandable narrative elements on our English-based dataset, the generalizability to other datasets, domains, languages, and more complex narratives is yet subject to more research. Large-scale and rigorous experiments (e.g., a linguistic evaluation of the constructs by experts) would be required to further validate AMR graphs' explainability, effectiveness, reliability, and accuracy in narrative extraction. 

\paragraph{Ethical Considerations.}
As conspiracy theories are a sensitive societal topic, we outline three primary ethical considerations of our research. 
First, our analyses are based on a publicly available dataset that includes information from publicly available news media. The presented results are highly aggregated and do not allow the identification of any individual website or person. The harm to human subjects is thus negligible. Second, as we leverage pretrained language models, we are also subject to their inherent biases. Third, our approach aims to better understand conspiracy narratives rather than advocating any of the knowledge attained. A better understanding could counteract conspiracy theories, but coincidentally also enable a better framing of conspiracy theories. Still, an improved understanding of diverging/competing frames in conspiracy and mainstream media can, in general, be seen as having a positive impact on society.

\section{Conclusion}
In the present work, we discussed how semantics derived from AMR graphs relate to the framing of narrative content. We showed that AMR is an ideal fit to analyze narrative frames, as we can directly extract context, characters, and plot from its graph representation. Using AMR, we introduced a conceptually simple to employ but flexible approach (\emph{C1}). We demonstrated the merits of our approach for framing analysis by contrasting conspiracy to mainstream media on three health-related topics (\emph{C2}), i.e., COVID-19, diseases, and pharmacology.

We observe that all three topics paint a similar picture of conspiracy media (i.e., a tendency towards beliefs instead of science). Hence, our approach provides a more holistic view of conspiracy narratives than previous research. 
We hope that our work inspires future research related to nuanced framing analysis.

\clearpage
\theendnotes
\clearpage

\bibliographystyle{apacite}
\bibliography{bib, anthology}

\newpage
\section{Supplemental Material}

\begin{figure*}[ht]
    \centering
    \captionsetup{type=figure}
    \includegraphics[width=0.99\columnwidth, decodearray={.1 .5 .5 .5 .5 .5}]{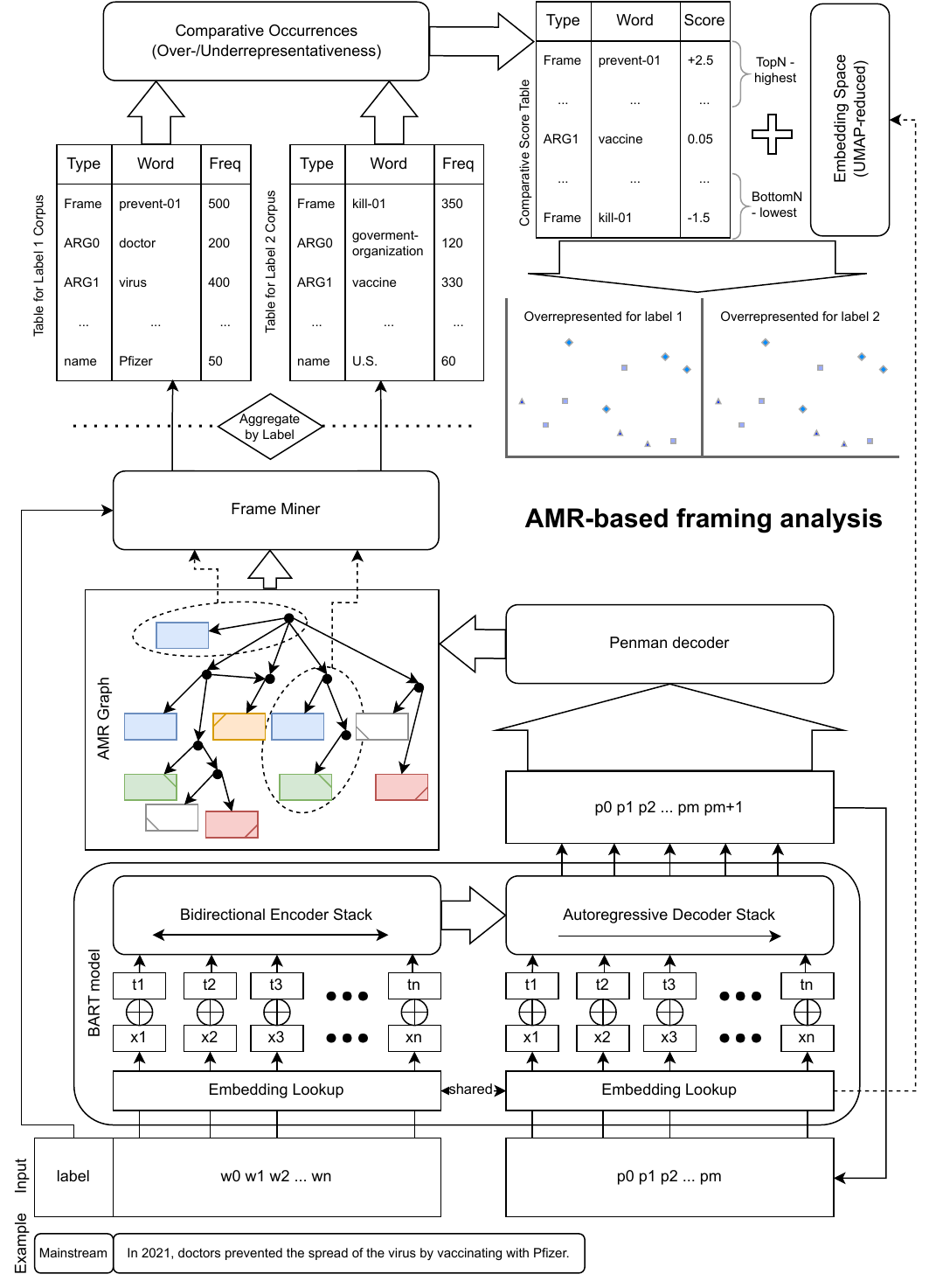}
    \caption{AMR-based Framing Analysis Approach Overview}
    \label{fig:framework}
\end{figure*}

\begin{table}[ht]
\begin{minipage}{0.99\textwidth}

    \centering
    \resizebox{.87\columnwidth}{!}{
    \rotatebox{90}{
    \begin{tabular}{ll|ll|ll|ll|ll|ll}
\toprule
         &  & \multicolumn{2}{c|}{Plot} & \multicolumn{2}{c|}{Characters} & \multicolumn{2}{c|}{Setting} & \multicolumn{2}{c|}{Moral of the Story} & \multicolumn{2}{c}{Entities} \\
         &  &          Conspiracy &        Mainstream &               Conspiracy &          Mainstream &          Conspiracy &          Mainstream &          Conspiracy &           Mainstream &           Conspiracy &              Mainstream \\
\midrule
COVID-19 & 1  &     destroy (24.91) &       say (43.95) &                i (39.65) &     case-04 (29.38) &        ever (27.06) &       early (11.73) &   counter-01 (9.03) &     ensure-01 (7.48) &    "America" (35.88) &      "COVID-19" (47.36) \\
         & 2  &       claim (23.10) &      test (29.09) &             book (29.93) &     test-01 (28.29) &         now (17.16) &    pandemic (10.15) &     alarm-01 (8.62) &   information (7.21) &      "Gates" (34.28) &      "Covid-19" (23.09) \\
         & 3  &        free (20.90) &   develop (21.72) &         military (25.37) &     vaccine (26.00) &      planet (14.88) &        week (10.14) &      free-01 (6.07) &       test-01 (6.83) &        "God" (28.94) &    "SARS-CoV-2" (21.57) \\
         & 4  &       state (20.59) &    infect (21.17) &          article (24.81) &     patient (24.83) &       today (13.47) &           by (9.25) &   keep-up-05 (5.97) &      treat-03 (6.83) &      "Wuhan" (28.81) &    "University" (19.35) \\
         & 5  &     believe (20.14) &  possible (20.98) &      amr-unknown (23.68) &     disease (24.56) &       above (12.25) &  date-entity (8.95) &         news (5.14) &       help-01 (5.95) &       "Bill" (28.22) &        "Health" (18.17) \\
         & 6  &        true (19.88) &      work (18.38) &              you (23.17) &     risk-01 (22.47) &        book (12.15) &     hospital (8.93) &   attempt-01 (5.01) &     reduce-01 (4.43) &         "US" (24.03) &  " coronavirus" (16.85) \\
         & 7  &         lie (18.90) &    reopen (17.32) &              law (22.56) &   infect-01 (21.64) &       below (11.67) &        after (8.68) &   benefit-01 (4.52) &     assess-01 (3.85) &     "Israel" (23.92) &        "Oxford" (14.73) \\
         & 8  &         war (18.86) &      case (16.54) &            media (21.88) &     company (21.43) &       world (11.49) &         home (8.35) &  possible-01 (4.50) &      allow-01 (3.84) &      "China" (21.81) &       "Zealand" (13.30) \\
         & 9  &  mainstream (18.55) &      wear (16.28) &            truth (21.42) &     symptom (20.21) &       final (11.30) &        month (8.10) &    battle-01 (4.45) &     advise-01 (3.80) &      "Earth" (19.88) &        "Brazil" (12.72) \\
         & 10 &       legal (17.90) &      risk (16.20) &  political-party (21.35) &        drug (19.70) &        just (10.46) &         city (8.07) &      save-02 (4.44) &       slow-01 (3.69) &      "Trump" (19.84) &       "Moderna" (12.29) \\
         & 11 &       cause (17.18) &      care (15.91) &            world (19.66) &       virus (18.30) &  publication (9.94) &        as-of (7.24) &      purpose (4.26) &        see-01 (3.48) &        "Big" (18.91) &           "and" (11.91) \\
         & 12 &      oppose (16.70) &       new (15.61) &            power (19.32) &        mask (18.23) &     tomorrow (9.90) &    community (7.23) &    resist-01 (4.22) &  organization (3.28) &  "Christian" (18.78) &           "NHS" (11.38) \\
         & 13 &        note (16.50) &    spread (15.59) &           war-01 (19.26) &    antibody (17.48) &       former (9.39) &      surface (6.97) &  suppress-01 (4.12) &    support-01 (3.18) &       "Iran" (18.76) &       "England" (11.19) \\
         & 14 &     natural (16.45) &      have (15.58) &             they (19.10) &     care-03 (17.08) &         list (8.97) &       so-far (6.68) &   destroy-01 (4.05) &     travel-01 (3.07) &      "Bible" (18.35) &            "UK" (10.57) \\
         & 15 &        word (16.39) &   contact (15.37) &          nothing (19.04) &  contact-01 (16.62) &         here (8.89) &      setting (6.59) &  distract-01 (3.99) &      check-01 (3.04) &       "West" (17.90) &   "AstraZeneca" (10.36) \\
\midrule
Disease & 1  &      govern (28.01) &         say (49.44) &                      you (43.43) &      disease (80.28) &      ever (26.35) &       country (16.90) &       alarm-01 (7.38) &    treat-03 (11.21) &        "US" (36.60) &           "HIV" (63.26) \\
        & 2  &       claim (27.98) &      infect (47.20) &  government-organization (37.47) &    infect-01 (40.25) &      here (20.92) &         early (15.86) &        free-01 (7.23) &   prevent-01 (8.93) &   "America" (32.58) &         "Ebola" (49.01) \\
        & 3  &        true (25.36) &       treat (35.50) &                 military (30.23) &        virus (38.44) &       now (18.49) &         after (14.60) &     counter-01 (6.31) &     fight-01 (6.30) &    "Israel" (30.29) &          "Zika" (47.78) \\
        & 4  &       state (25.32) &    transmit (31.94) &                        i (29.88) &     treat-03 (35.35) &    planet (17.88) &   outbreak-29 (13.49) &      effort-01 (5.55) &    ensure-01 (6.24) &    "Russia" (30.14) &        "cancer" (39.39) \\
        & 5  &     natural (25.06) &      spread (30.48) &                    media (26.96) &  outbreak-29 (33.95) &     below (16.17) &          area (12.38) &      profit-01 (5.53) &  information (6.05) &       "God" (24.22) &          "AIDS" (27.75) \\
        & 6  &    engineer (23.70) &    diagnose (26.98) &                  article (26.71) &      case-04 (33.57) &    former (14.31) &  world-region (12.28) &        purpose (5.40) &    assess-01 (5.85) &     "Earth" (24.21) &         "Congo" (26.01) \\
        & 7  &      actual (23.19) &        risk (25.07) &              amr-unknown (26.68) &      risk-01 (31.23) &     above (14.23) &        region (11.82) &        save-02 (5.40) &    reduce-01 (5.65) &     "Trump" (23.01) &      "HIV/AIDS" (25.46) \\
        & 8  &       legal (23.01) &  impregnate (23.84) &                     book (25.71) &        woman (29.97) &     world (13.63) &     pregnancy (10.37) &     justify-01 (5.39) &      help-01 (4.88) &       "Jew" (22.44) &        "Health" (24.94) \\
        & 9  &     destroy (22.19) &     prevent (23.63) &                     they (25.33) &         drug (28.02) &  all-over (11.31) &          year (10.14) &     benefit-01 (4.95) &   address-02 (4.82) &     "China" (21.80) &           "WHO" (23.82) \\
        & 10 &  mainstream (21.16) &    outbreak (23.26) &                    truth (23.57) &  transmit-01 (27.80) &   article (10.90) &           east (9.95) &      create-01 (4.88) &    screen-01 (4.78) &      "Iraq" (21.63) &        "Z Zika" (21.38) \\
        & 11 &         lie (20.86) &        care (23.14) &                   weapon (22.89) &      patient (27.74) &       war (10.65) &          since (9.51) &  experiment-01 (4.87) &     check-01 (4.67) &     "Gates" (20.82) &        "Africa" (19.98) \\
        & 12 &      create (20.72) &      detect (22.88) &                      law (22.08) &          she (26.49) &      book (10.62) &          stage (9.50) &           seat (4.74) &   respond-01 (4.49) &  "Monsanto" (20.80) &        "Uganda" (18.89) \\
        & 13 &      oppose (20.43) &        case (22.66) &                    power (21.90) &     mosquito (25.80) &     today (10.16) &        current (9.16) &      expand-01 (4.67) &    combat-01 (4.38) &       "War" (20.75) &  "Organization" (17.70) \\
        & 14 &        read (20.19) &         new (22.41) &                      oil (21.62) &       health (25.37) &    behind (10.01) &           city (8.10) &       prove-01 (4.66) &   support-01 (4.33) &     "Syria" (20.50) &           "DRC" (17.47) \\
        & 15 &     believe (20.19) &      travel (22.32) &                   war-01 (21.30) &         cell (25.03) &   facility (9.91) &          month (7.94) &       force-01 (4.61) &      care-03 (4.27) &      "Iran" (19.55) &        "Dengue" (16.10) \\
\midrule
Pharma & 1  &     destroy (33.30) &          say (71.64) &                      you (58.11) &       company (53.50) &        ever (34.17) &         year (23.63) &      profit-01 (9.52) &      disease (9.50) &  "America" (44.91) &     "COVID-19" (39.24) \\
       & 2  &     natural (28.75) &      develop (45.22) &                        i (38.97) &          drug (52.56) &         now (20.76) &  date-entity (17.03) &       alarm-01 (7.13) &    ensure-01 (8.98) &      "Big" (35.03) &           "UK" (23.26) \\
       & 3  &  mainstream (28.57) &        price (37.93) &                    media (31.46) &       vaccine (37.57) &       below (19.15) &         week (16.19) &           news (7.06) &     treat-03 (7.90) &      "God" (32.06) &        "India" (23.20) \\
       & 4  &      govern (26.98) &         cost (26.99) &  government-organization (30.76) &      price-01 (36.62) &        here (18.45) &        month (16.12) &        free-01 (6.70) &   develop-02 (7.40) &    "Gates" (30.30) &     "Covid-19" (21.92) \\
       & 5  &        know (26.79) &          new (25.10) &                    truth (30.75) &       disease (30.94) &      planet (16.44) &           by (13.70) &     destroy-01 (6.57) &      test-01 (6.80) &      "CDC" (27.00) &          "NHS" (21.81) \\
       & 6  &        true (26.55) &       infect (23.70) &                  article (30.21) &       cost-01 (29.57) &       above (15.82) &         city (10.91) &       right-05 (5.66) &  information (5.98) &     "Bush" (26.98) &   "University" (21.73) \\
       & 7  &         lie (25.39) &     possible (21.89) &                     they (29.98) &    develop-02 (26.54) &       today (14.25) &     pandemic (10.82) &        save-02 (5.15) &      help-01 (5.82) &     "Bill" (26.41) &       "Pfizer" (21.67) \\
       & 8  &        free (24.79) &         have (20.15) &              amr-unknown (29.50) &       patient (23.93) &  url-entity (13.70) &        early (10.57) &      defend-01 (5.13) &   address-02 (5.55) &   "Russia" (23.84) &      "Moderna" (20.98) \\
       & 9  &        read (23.66) &          low (19.72) &              corporation (29.00) &        try-01 (23.70) &      behind (12.70) &       country (9.95) &     benefit-01 (5.02) &      meet-01 (5.04) &     "Iraq" (22.77) &      "measles" (20.54) \\
       & 10 &     corrupt (23.49) &        treat (19.61) &                 military (28.80) &       protein (23.00) &       realm (10.38) &         after (9.05) &  experiment-01 (4.98) &   improve-01 (4.71) &    "Obama" (22.74) &    "Wakefield" (20.49) \\
       & 11 &        kill (23.08) &         test (19.28) &                     book (28.78) &  vaccinate-01 (22.19) &     article (10.28) &    university (8.69) &       cause-01 (4.88) &      measles (4.56) &     "Iran" (22.64) &     "Medicare" (20.26) \\
       & 12 &     control (22.97) &       access (19.01) &                      man (24.60) &      antibody (21.78) &         war (10.03) &        so-far (8.47) &     justify-01 (4.70) &  immunize-01 (4.31) &    "Wuhan" (22.23) &       "Health" (20.12) \\
       & 13 &       cause (22.95) &     research (18.82) &                   war-01 (24.60) &      treat-03 (21.75) &         then (9.94) &     community (8.00) &        live-01 (4.68) &    assess-01 (4.21) &      "CIA" (21.48) &       "Oxford" (19.84) \\
       & 14 &    engineer (21.77) &         work (18.47) &                    power (22.68) &     infect-01 (21.70) &        world (9.83) &         globe (7.98) &        version (4.67) &       try-02 (4.05) &   "Pharma" (21.01) &  "AstraZeneca" (19.11) \\
       & 15 &         war (21.60) &  manufacture (18.22) &                  citizen (22.56) &     candidate (21.61) &     about-to (9.52) &         state (7.62) &     counter-01 (4.66) &      make-01 (4.00) &      "War" (20.92) &   "SARS-CoV-2" (18.22) \\
\bottomrule
\end{tabular}
}}
    \caption{Overrepresented elements.}
    \label{tab:my_label}

\end{minipage}
\end{table}

\end{document}